\documentclass{bmvc2k}
\usepackage{graphicx,subfigure}

\title{Can Super Resolution be used to improve Human Pose Estimation in Low Resolution Scenarios?}

\addauthor{Peter Hardy}{https://www.ecs.soton.ac.uk/people/ptdh1c20}{1}
\addauthor{Srinandan Dasmahapatra}{https://www.ecs.soton.ac.uk/people/srinanda}{1}
\addauthor{Hansung Kim}{https://www.ecs.soton.ac.uk/people/hk1f20}{1}

\addinstitution{
 Vision, Learning and Control Research Group\\
 University of Southampton\\
 Hampshire, UK
}

\runninghead{P. Hardy et al}{Super Resolution in Human Pose Estimation}

\begin{document}

\maketitle

\begin{abstract}
The results obtained from state of the art human pose estimation (HPE) models degrade rapidly when evaluating people of a low resolution, but can super resolution (SR) be used to help mitigate this effect? By using various SR approaches we enhanced two low resolution datasets and evaluated the change in performance of both an object and keypoint detector as well as end-to-end HPE results. We remark the following
observations. {First we find that for people who were originally depicted at a low resolution (segmentation area in pixels), their keypoint detection performance would improve once SR was applied.} Second, the keypoint detection performance gained is dependent on that persons pixel count in the original image prior to any application of SR; keypoint detection performance was improved when
SR was applied to people with a small initial segmentation area, but degrades as this becomes larger. To address this we introduced a novel Mask-RCNN approach, utilising a segmentation area threshold to decide when to use SR during the keypoint detection step. {This approach achieved the best results on our low resolution datasets for each HPE performance metrics.}
\end{abstract}


\section{Introduction}

Human Pose Estimation (HPE) and keypoint detection are important research topics in computer vision, with many real-world applications such as action recognition and interactive media \cite{Luvizon2020MultitaskDL} \cite{review_single_image}. Although modern HPE models obtain impressive results on popular datasets such as COCO \cite{COCO} and MPII \cite{MPII}, their performance degrades substantially when evaluating people of a small scale and low resolution \cite{small_people_bad}. During keypoint detection, current HPE models utilise Convolutional Neural Networks (CNN). However, as convolutions have a limited robustness to an objects scale \cite{pmlr-v77-takahashi17a}, ongoing work creating scale invariant CNN architectures remains a key research focus \cite{NIPS2010_01f78be6} \cite{Sun2019DeepHR} \cite{scale_invariance} \cite{9052469}. In contrast, little research exploring how a persons' resolution can be improved for HPE has been undertaken. Super resolution (SR) has been touted within object detection as a panacea for issues of image quality \cite{SR_OBJECT_DETECTION} \cite{8622135} \cite{SR_OBJ}, but could it also benefit HPE? This paper will explore multiple SR techniques on low resolution imagery to address this question. By evaluating the performance of a HPE model at different stages of the end-to-end process, we will establish the effect that SR has on HPE and how it varies depending on the target persons' initial resolution.


\section{Background}

\subsection{Human Pose Estimation}

The objective of HPE is to locate and group specific keypoints or bodyparts (shoulder, ankle, etc) from a given image in order to create a human pose \cite{Sun2019DeepHR}. Current HPE methods fall into two categories: bottom-up and top-down approaches. Bottom-up approaches only consist of a keypoint detector, which is used to detect all keypoints in a given image. They then use a grouping algorithm, or human body fitting model, to associate the keypoints with each other in order to create a human pose. By comparison, top-down approaches utilise both an object and keypoint detection component. They start by first detecting the bounding box of each person in an image and then perform keypoint detection inside each bounding box. This negates the need for a grouping algorithm as all the keypoints in each bounding box are assumed to correspond to the same person \cite{CHEN2020102897}. As the number of people in a given scene increases, so does the computational cost of top-down approaches. However, this approach is more accurate overall as more people within a scene are detected. {The most popular approach to keypoint detection \cite{cascadingpyramid} \cite{hourglass} \cite{8237406} \cite{Sun2019DeepHR} \cite{cao2019openpose} \cite{deeppose} \cite{Cheng_2020_CVPR} \cite{Chen_2018_CVPR} is known as heatmap regression \cite{heatmapreg}. A CNN head extracts the initial features and reduces the resolution of the input image, this is followed by a main body which outputs feature maps with the same size as the input feature map, and is followed by a regressor which estimates the predicted heatmaps \cite{Sun2019DeepHR}. Ground truth heatmaps are constructed by using a 2D Gaussian kernel on the given ground truth keypoint and are used to supervise the predicted heatmaps by reducing the L2 loss.} Since its introduction by Tompson et al. \cite{first_heatmap}, heatmaps have become the default method for keypoint detection due to its ease of implementation and much higher accuracy than traditional coordinate regression \cite{deeppose} \cite{hourglass}.

\subsection{Super Resolution}

Currently used in multiple real-world applications such as security \cite{ZHANG2010848} \cite{10.1007/978-3-319-41778-3_18} and medical imaging \cite{medical} \cite{medical_2} \cite{sr_survey}, SR refers to the process of recovering accurate high resolution images from their low resolution counterparts. Modern state of the art SR performance is obtained from deep learning approaches, such as generative adversarial networks (GAN) \cite{ESRGAN} \cite{srgan} and auto-encoder architectures \cite{USRNET} \cite{9025499}. While there exists many different ways of assessing the performance of SR models (structural similarity index \cite{SSIM}, feature similarity index \cite{FSIM}, etc) the most commonly used metric is the peak signal-to-noise ratio (PSNR). Although PSNR is regularly used as a training metric, the output images generated by maximising PSNR correlates poorly with image quality as perceived by the human eye \cite{PSNR_BAD} \cite{psnr_still_bad}. This disparity is surprising when findings of recent studies have improved the overall object detection performance in low resolution imagery when combining SR with an object detector \cite{SR_OBJECT_DETECTION} \cite{8622135} \cite{SR_OBJ}. Providing more evidence that deep learning approaches may not perceive image quality the same way as humans, and may in fact learn completely different associations when identifying objects. Some studies however have found a negative impact on object detection performance due to SR if the resolution of the object in the original image is extremely low \cite{super_resolution_bad_planes}.


\section{Method}

While it is difficult to define low resolution with a numerical value, intuition tells us that a low resolution image will be more pixelated and less informative than a high resolution one. We can therefore infer that commonly used computer vision datasets are not low resolution, due to the clarity of the images present. In order to evaluate if SR can improve the HPE results of low resolution people, we used bicubic downsampling to create two low resolution versions ($\frac{1}{2}$ and $\frac{1}{4}$ scale) of the COCO validation dataset. We then applied various SR techniques on these low resolution datasets to increase their resolution by a factor of 4. This would then allow us to compare the HPE results between the low resolution images and their SR counterparts. The COCO dataset was chosen for this study as each images annotation also contains the segmentation area (in pixels) of each person in an image. This allowed us to investigate how the effects of SR differ depending on the persons starting segmentation area, as SR may have an adverse effect the lower the initial segmentation area due to the limited amount of starting pixels to reconstruct a high resolution person from. The SR approaches we used to enhance our images were standard bicubic interpolation, ESRGAN \cite{ESRGAN} and USRNET \cite{USRNET}. For ESRGAN and USRNET we used a GAN version of each model (ESRGAN and USRGAN), and a PSNR maximising version of each model (ESRNET and USRNET). For our HPE model we used HRNET \cite{Sun2019DeepHR}, a top-down based approach which achieved one of the highest accuracies across various keypoint datasets at the time of writing. We chose a top-down approach for this study as it consists of both an object and keypoint detection component, which allowed us to test the possible effects that SR has at multiple stages of the end-to-end HPE process.

\subsection{Object Detection with Super Resolution}

The object detector that we used was Faster R-CNN with a resnet-101 and feature pyramid network backbone \cite{RCNNFPN}. This was trained on the standard unaltered version of the COCO training dataset and {the AP and AR performance of this model on the default COCO validation dataset can be seen in Table \ref{original_detection_results}}. As the COCO annotation groups people into a small, medium and large subgroup (S, M and L) depending on that persons given segmentation area, a persons subgroup would usually change when SR is applied. In order to compare the effect that SR has on each subgroup fairly, if someone was defined as small, medium or large in the scaled down image, they would also be defined as that subgroup in the results of the SR image. {It is for this reason we have not reported the S, M, or L results of the default COCO dataset as no fair comparison can be made.} The results showing the average precision (AP) and average recall (AR) of our detector on the low resolution datasets and their SR counterparts can be seen in Table \ref{Detection_Results_Scale2} and \ref{Detection_Results_Scale4}

\begin{table}[!htb]
\centering
\begin{tabular}{|c|cc|}
\hline
Dataset &
  AP &
  AR \\ \hline
COCO & 0.545   & 0.612  \\
\hline
\end{tabular}
\vspace{+1mm}
\caption{The person detection results on the default COCO dataset}
\label{original_detection_results}
\end{table}

\begin{table}[!htb]
\centering
\begin{tabular}{|c|clllclcc|}
\hline
Dataset &
  AP &
  $\text{AP}_\text{S}$ &
  $\text{AP}_\text{M}$ &
  $\text{AP}_\text{L}$ &
  AR &
  $\text{AR}_\text{S}$ &
  $\text{AR}_\text{M}$ &
  $\text{AR}_\text{L}$ \\ \hline
LR ($\frac{1}{2}$ scale) & 0.507          & 0.394 & 0.683 & 0.752          & 0.571 & 0.459 & 0.741          & 0.829 \\
Bicubic                  & 0.511          & 0.399 & 0.684 & 0.752          & 0.577 & 0.467 & 0.742          & 0.828 \\
ESRGAN \cite{ESRGAN}                   & 0.515          & 0.402 & 0.684 & 0.755          & 0.581 & 0.473 & 0.743          & 0.829 \\
ESRNET \cite{ESRGAN}                   & 0.521          & 0.409 & 0.692 & 0.752          & 0.589 & 0.483 & \textbf{0.749} & 0.832 \\
USRGAN \cite{USRNET}                   & \textbf{0.522} & 0.410 & 0.689 & \textbf{0.756} & 0.588 & 0.481 & \textbf{0.749} & 0.832 \\
USRNET \cite{USRNET} &
  \textbf{0.522} &
  \textbf{0.411} &
  \textbf{0.690} &
  0.754 &
  \textbf{0.590} &
  \textbf{0.485} &
  0.748 &
  \textbf{0.834} \\ \hline
\end{tabular}
\vspace{+1mm}
\caption{The person detection results of the $\frac{1}{2}$ scale low resolution (LR) dataset and the SR datasets obtained by upscaling the LR dataset by a factor of 4. The best approach for each evaluation metric is highlighted in bold.}
\label{Detection_Results_Scale2}
\end{table}

\begin{table}[!htb]
\centering
\begin{tabular}{|c|clllclcc|}
\hline
Dataset &
  AP &
  $\text{AP}_\text{S}$ &
  $\text{AP}_\text{M}$ &
  $\text{AP}_\text{L}$ &
  AR &
  $\text{AR}_\text{S}$ &
  $\text{AR}_\text{M}$ &
  $\text{AR}_\text{L}$ \\ \hline
LR ($\frac{1}{4}$ scale) & 0.387 & 0.322 & 0.688 & 0.716          & 0.448 & 0.378 & 0.762          & 0.844          \\
Bicubic                                   & 0.413 & 0.351 & 0.697 & 0.736          & 0.478 & 0.414 & 0.768          & 0.868          \\
ESRGAN \cite{ESRGAN}                                    & 0.445 & 0.385 & 0.721 & 0.739          & 0.509 & 0.448 & 0.786          & \textbf{0.891} \\
ESRNET \cite{ESRGAN}                                    & 0.454 & 0.394 & 0.728 & \textbf{0.749} & 0.519 & 0.459 & \textbf{0.793} & 0.886          \\
USRGAN \cite{USRNET} &
  \textbf{0.456} &
  \textbf{0.396} &
  \textbf{0.729} &
  0.743 &
  \textbf{0.521} &
  \textbf{0.461} &
  \textbf{0.793} &
  0.874 \\
USRNET \cite{USRNET}                                    & 0.452 & 0.392 & 0.725 & 0.736          & 0.518 & 0.458 & 0.790          & 0.870          \\ \hline
\end{tabular}
\vspace{+1mm}
\caption{The person detection results of the $\frac{1}{4}$ scale low resolution (LR) dataset and the SR images obtained by upscaling the LR dataset by a factor of 4. The best approach for each evaluation metric is highlighted in bold.}
\label{Detection_Results_Scale4}
\end{table}

\begin{figure}[!htb]

\centering
\includegraphics[width=.5\textwidth]{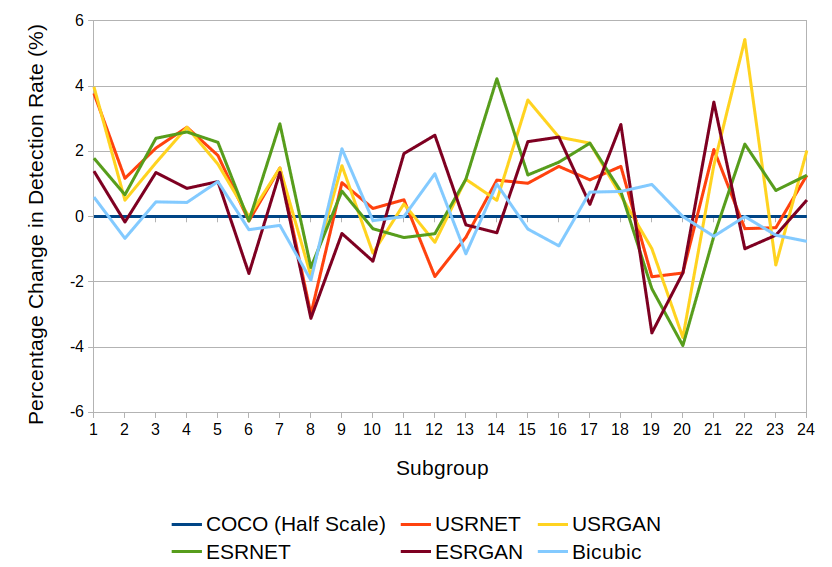}\hfill
\includegraphics[width=.5\textwidth]{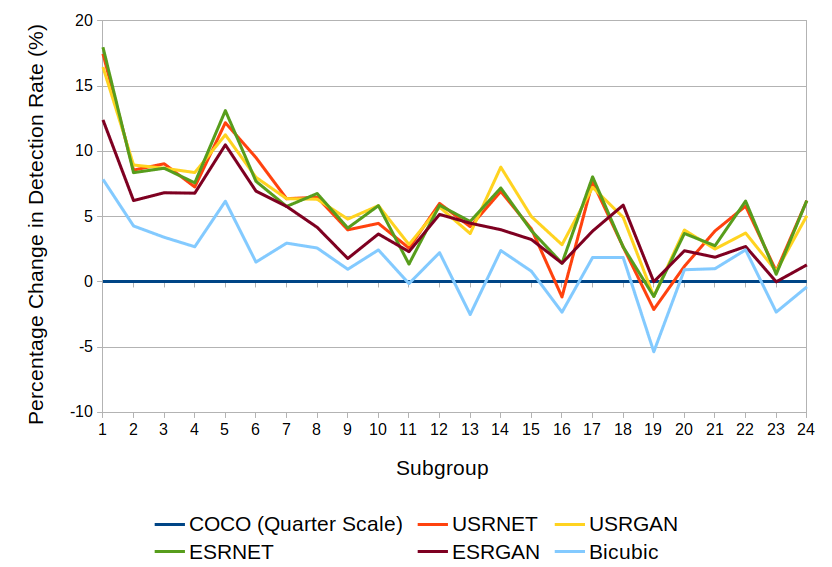}\hfill

\caption{The percentage change in detection rate due to SR for each of our 24 subgroups created from $\frac{1}{2}$ scale (left panel) and $\frac{1}{4}$ scale (right panel) datasets and their SR counterparts.}
\label{fig:detection_across_res}
\end{figure}

Our results show that the overall performance (AP and AR) of the object detector improved once SR was applied. This concurs with previous studies in this area \cite{SR_OBJ} \cite{SR_OBJECT_DETECTION} \cite{8622135}. One recent study however found that the lower the original resolution of the object we are detecting, the worse the object detector would perform after SR was applied \cite{super_resolution_bad_planes}. As the small subgroup contained people with a segmentation area between 1 and $32^2$ pixels, we could not confirm whether all people of a smaller segmentation area had improved, simply that this group as a whole did. In order to determine if the improvements in detection rate were skewed by performance variations in subgroups, we conducted a further test. We created 24 new subgroups from our data, grouping people of a similar segmentation area together. The segmentation areas of people within each subgroup (1-24) for our $\frac{1}{2}$ scale dataset commenced at 1-500 and concluded at 11501-12000 increasing by 500 for each subgroup. For the $\frac{1}{4}$ scale dataset the segmentation areas of people within each subgroup increased by 125, starting at 1-125 and concluding at 2876-3000. We then evaluated the performance of the object detector across these 24 subgroups. Our findings can be seen in Figure \ref{fig:detection_across_res} which shows the percentage increase or decrease in object detection rate for each subgroup once SR had been applied.

Contradicting with previous findings, our results demonstrate that the lower the original segmentation area of the object we wish to detect, the more likely it will be detected once SR is performed. Additionally, we found what seems to be a soft threshold of 3000 pixels, as shown by subgroup 6 in the left panel (segmentation area of 2501-3000). People who had an initial segmentation area below this threshold had their detection rate improved once SR was performed on them, aside from bicubic interpolation and ESRGAN for subgroup 2. For people above this threshold however, it is not clear if the overall detection performance would improve or worsen once SR was applied as the results seem to be sporadic in nature. In the right panel the largest subgroup contained people with a segmentation area of between 2876-3000, and as we decrease the subgroup number, and therefore initial segmentation area, there is a gradual improvement in object detection rate.


\subsection{Keypoint Detection with Super Resolution}

This section will examine how SR affects the keypoint detection component of the end-to-end HPE process. Our HRNET \cite{Sun2019DeepHR} model was trained on the standard COCO training dataset and {the AP and AR performance when evaluating ground truth bounding boxes in the original COCO dataset can be seen in Table \ref{og_keypoint_results}}. As our previous results have shown that the lower the resolution of your object, the better the object detection rate will be once SR had been performed. We now wanted to determine if this also held true for keypoint detection. To eliminate object detection as a variable, we provided HRNET with the ground truth bounding boxes of people in each image. This allowed us to analyse the overall effect that SR has on keypoint detection in low resolution imagery, given that the object detection results are identical. The evaluation metric we used for this study is based on Object Keypoint Similarity (OKS): 

\begin{equation}
\frac{\Sigma_i \text{ exp}(-d_i^2/2s^2k^2_i)\delta(v_i > 0)}{\Sigma_i \delta(v_i > 0)},
\end{equation}
where $d_i$ is the Euclidean distance between the detected and corresponding ground truth keypoint, $v_i$ is the visibility flag of the ground truth keypoint, $s$ is the objects scale and $k_i$ is a per-keypoint constant that controls falloff. In our results we report standard average precision and recall scores \cite{COCO}: AP, the mean of the AP scores at 10 positions (OKS = 0.50, 0.55, ..., 0.90, 0.95), $\text{AP}_\text{M}$ for medium objects, $\text{AP}_\text{L}$ for large objects, AR (the mean of AR scores OKS = 0.50, 0.55, ..., 0.90, 0.95) and AR for medium and large people ($\text{AR}_\text{M}$ and $\text{AR}_\text{L}$ respectively). {$\text{AP}_\text{S}$ is not reported during keypoint detection as people with a segmentation area < $32^2$ do not have their keypoints annotated in the COCO dataset.} The results of our keypoint detector on the low and SR datasets can be seen in Tables \ref{Keypoint_Results_Scale2} and \ref{Keypoint_Results_Scale4}.

\begin{table}[!htb]
\centering
\begin{tabular}{|c|cc|}
\hline
Dataset & AP   & AR  \\ \hline
COCO & 0.765  & 0.793      \\ \hline
\end{tabular}
\vspace{+1mm}
\caption{The AP and AR performance of HRNET \cite{Sun2019DeepHR} on the default COCO dataset when evaluating ground truth bounding boxes.}
\label{og_keypoint_results}
\end{table}

\begin{table}[!htb]
\centering
\begin{tabular}{|c|cllccc|}
\hline
Dataset                  & AP    & $\text{AP}_\text{M}$ & $\text{AP}_\text{L}$ & AR    & $\text{AR}_\text{M}$ & $\text{AR}_\text{L}$ \\ \hline
COCO $\frac{1}{2}$ Scale & 0.722 & 0.765           & \textbf{0.841}                & 0.752 & 0.794                & \textbf{0.880}                \\
Bicubic & 0.728 & 0.763 & 0.835 & 0.760 & 0.764 & 0.875 \\
ESRGAN \cite{ESRGAN}  & 0.729 & 0.764 & 0.825 & 0.761 & 0.796 & 0.866 \\
ESRNET \cite{ESRGAN} & \textbf{0.744} & \textbf{0.774} & 0.831 & \textbf{0.773} & \textbf{0.803} & 0.873 \\
USRGAN \cite{USRNET} & 0.735 & 0.769 & 0.826 & 0.766 & 0.798 & 0.870 \\
USRNET \cite{USRNET} & 0.741 & 0.772 & 0.832 & 0.772 & 0.802 & 0.873 \\ \hline
\end{tabular}
\vspace{+1mm}
\caption{The performance of HRNET \cite{Sun2019DeepHR} on the $\frac{1}{2}$ scale dataset and that same dataset upscaled by a factor of 4 using the various SR techniques. The best result for each evaluation metric is higlighted in bold.}
\label{Keypoint_Results_Scale2}
\end{table}

\begin{table}[!htb]
\centering
\begin{tabular}{|c|cllccc|}
\hline
Dataset & AP             & $\text{AP}_\text{M}$ & $\text{AP}_\text{L}$ & AR             & $\text{AR}_\text{M}$ & $\text{AR}_\text{L}$ \\ \hline
COCO $\frac{1}{4}$ Scale & 0.538 & 0.791 & 0.800          & 0.573 & 0.830 & 0.879 \\
Bicubic                  & 0.601 & 0.796 & \textbf{0.801} & 0.637 & 0.836 & 0.882 \\
ESRGAN \cite{ESRGAN}                   & 0.627 & 0.810 & 0.786          & 0.664 & 0.845 & 0.875 \\
ESRNET \cite{ESRGAN} & \textbf{0.649} & \textbf{0.813}  & \textbf{0.801}       & \textbf{0.684} & 0.848                & \textbf{0.888}       \\
USRGAN \cite{USRNET}                  & 0.635 & 0.810 & 0.794          & 0.670 & 0.846 & 0.882 \\
USRNET \cite{USRNET} & 0.647          & \textbf{0.813}  & 0.795                & 0.681          & \textbf{0.849}       & \textbf{0.888}       \\ \hline
\end{tabular}
\vspace{+1mm}
\caption{The performance of HRNET \cite{Sun2019DeepHR} on the $\frac{1}{4}$ scale dataset and that same dataset upscaled by a factor of 4 using the various SR techniques. The best result for each evaluation metric is higlighted in bold.}
\label{Keypoint_Results_Scale4}
\end{table}

\begin{figure}[!htb]

\centering
\includegraphics[width=.5\textwidth]{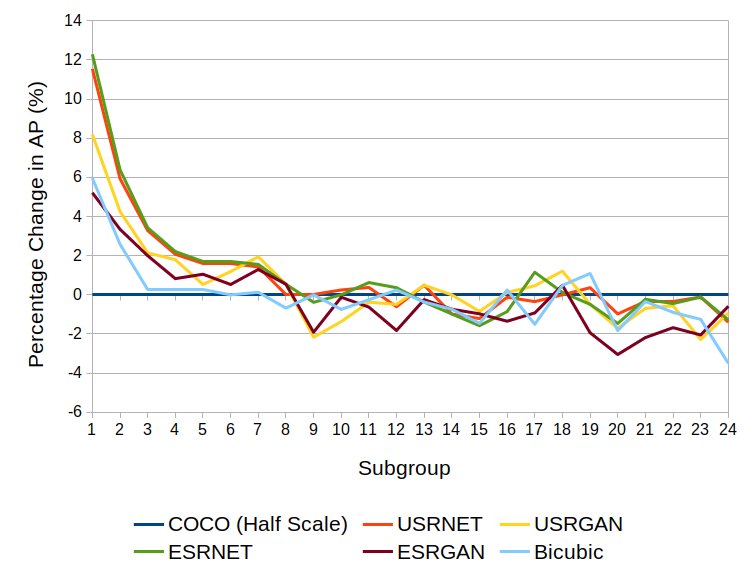}\hfill
\includegraphics[width=.5\textwidth]{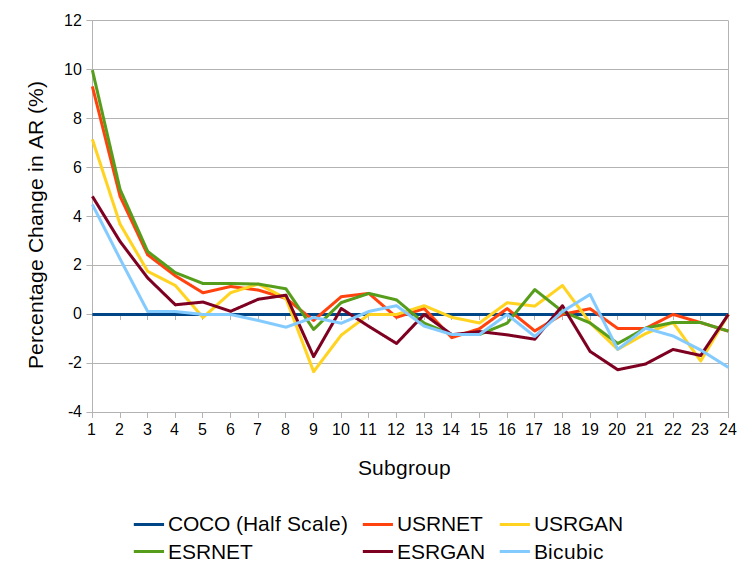}\hfill

\caption{The percentage change in AP (left panel) and AR (right panel) due to SR across our 24 subgroups of the $\frac{1}{2}$ scale dataset and their SR counterparts.}
\label{fig:keypoint_across_res_scale2}
\end{figure}

\begin{figure}[!htb]

\centering
\includegraphics[width=.5\textwidth]{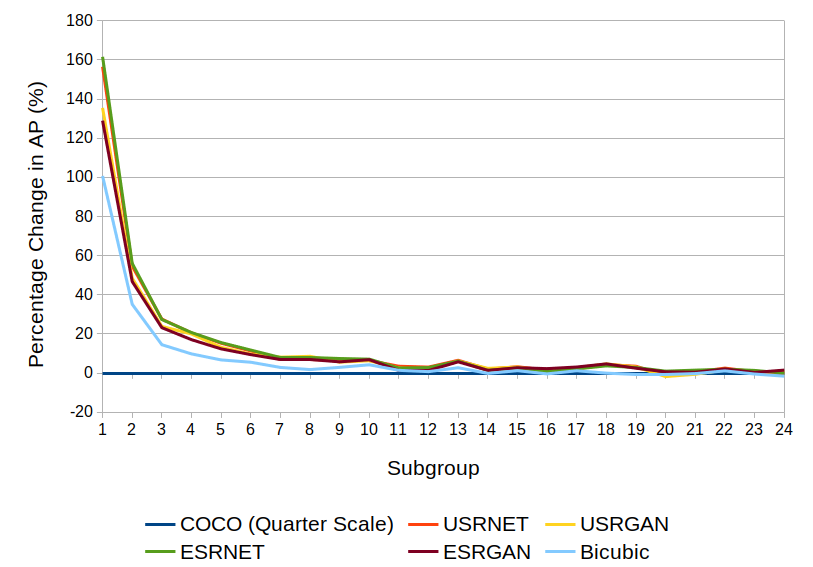}\hfill
\includegraphics[width=.5\textwidth]{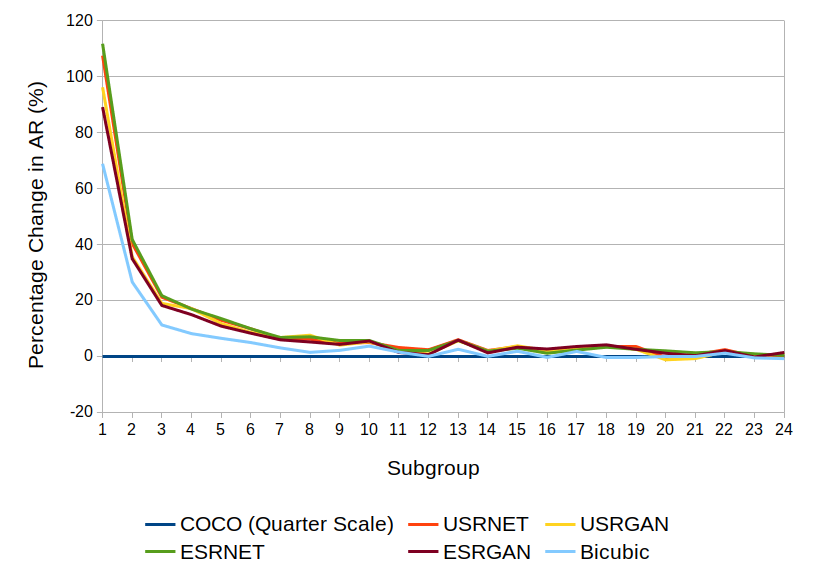}\hfill

\caption{The percentage change in AP (left panel) and AR (right panel) due to SR across our 24 subgroups of the $\frac{1}{4}$ scale dataset and their SR counterparts.}
\label{fig:keypoint_across_res_scale4}
\end{figure}

As our results show, the overall performance of our keypoint detector (AP and AR) improved when evaluating the SR versions of both the $\frac{1}{2}$ and $\frac{1}{4}$ scale dataset. When we look closely however, we can see that simply stating the performance would improve for all observations would be incorrect. Examining the keypoint detection performance for the large subgroup of people ($\text{AP}_\text{L}$ and $\text{AR}_\text{L}$) in the $\frac{1}{2}$ scale dataset, we can see that there was performance degradation as a result of SR. Additionally, not every SR approach we used improved the AP and AR of the medium subgroup of people ($\text{AP}_\text{M}$ and $\text{AR}_\text{M}$), as both bicubic interpolation and ESRGAN \cite{ESRGAN} actually made these results worse. Our findings seemed to hint at a person segmentation area threshold for keypoint detection, above which using SR on a person would worsen the performance of the keypoint detector. To confirm this, we evaluated the change in keypoint
detection performance across our 24 subgroups. From this we could determine if there is in-fact an upper limit in object segmentation area, above which the keypoint detection performance would worsen once SR was applied. The results are shown in Figure \ref{fig:keypoint_across_res_scale2} and \ref{fig:keypoint_across_res_scale4}.

Our figures show that as the initial segmentation area of the people we are evaluating increases, the benefits gained by applying SR for keypoint detection decreases. For our dataset, the threshold beyond which applying SR seems to have a negative affect on keypoint detection, is a segmentation area of between 3501-4000 (subgroup 8 in Figure \ref{fig:keypoint_across_res_scale2}). As beyond this value the percentage change in AP and AR once SR was applied starts to become negative. For people within our smallest subgroup however (subgroup 1 in Figure \ref{fig:keypoint_across_res_scale4}) it is worth noting just how prominent the performance increase is, as we observed a 160\% increase in AP performance and 110\% increase in AR performance.


\subsection{End-to-End Results}

\begin{figure}[!htb]

\centering
\subfigure[$\frac{1}{2}$ Scale Low Resolution Keypoints]{\label{fig:b}\includegraphics[width=.45\textwidth]{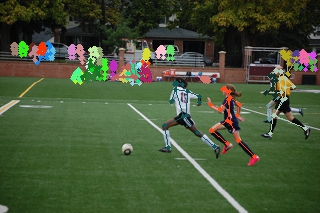}}
\subfigure[ESRNET SR of (a) Keypoints]{\label{fig:b}\includegraphics[width=.45\textwidth]{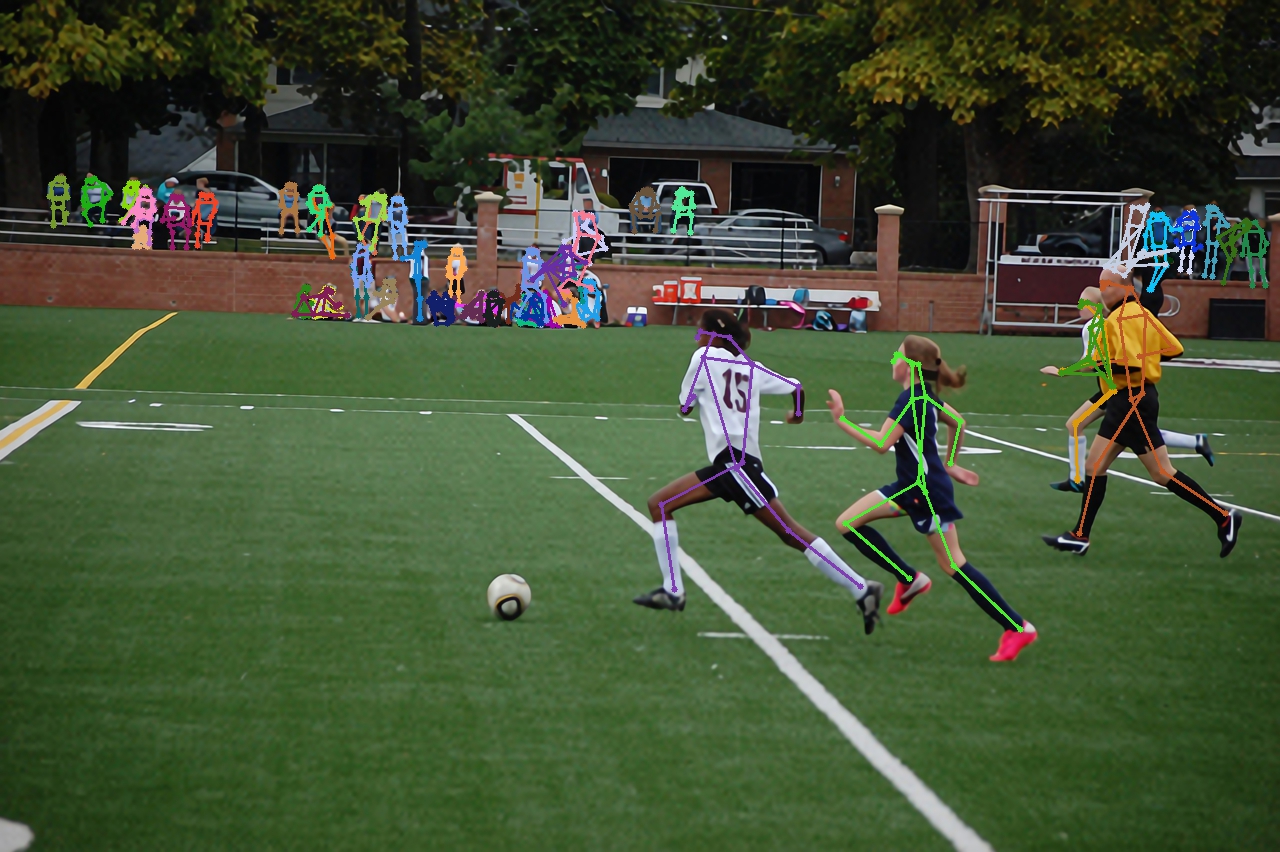}}
\subfigure[$\frac{1}{4}$ Scale Low Resolution Keypoints]{\label{fig:b}\includegraphics[width=.45\textwidth]{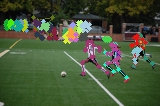}}
\subfigure[ESRNET SR of (c) Keypoints]{\label{fig:b}\includegraphics[width=.45\textwidth]{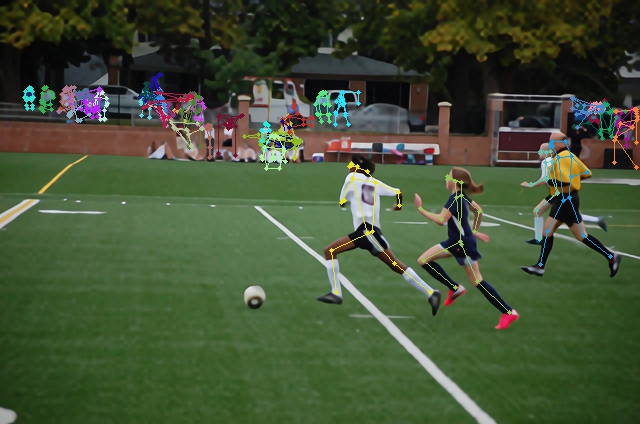}}

\caption{A visualisation of the keypoints detected in both a low resolution image and its SR counterpart.}
\label{fig:betterkeypoints}
\end{figure}

This section will evaluate the final keypoint detection performance obtained when using SR for the entire end-to-end top-down HPE process. The results of which can be seen in Table \ref{end_to_end_scale2} and \ref{end_to_end_scale4}. Additionally a visualisation showing the increase in HPE performance due to applying SR can be seen in Figure \ref{fig:betterkeypoints}

\begin{table}[!htb]
\centering
\begin{tabular}{|c|cllccc|}
\hline
Dataset                  & AP    & $\text{AP}_\text{M}$ & $\text{AP}_\text{L}$ & AR             & $\text{AR}_\text{M}$ & $\text{AR}_\text{L}$ \\ \hline
COCO $\frac{1}{2}$ Scale & 0.704 & 0.758           & \textbf{0.835}       & 0.747          & 0.799                & \textbf{0.879}       \\
Bicubic & 0.709          & 0.756 & 0.832 & 0.753 & 0.796 & 0.876 \\
ESRGAN  & 0.707          & 0.756 & 0.832 & 0.753 & 0.796 & 0.876 \\
ESRNET                   & 0.721 & \textbf{0.768}  & 0.827                & \textbf{0.766} & \textbf{0.805}       & 0.872                \\
USRGAN  & 0.715          & 0.761 & 0.827 & 0.760 & 0.801 & 0.874 \\
USRNET  & \textbf{0.722} & 0.766 & 0.828 & 0.766 & 0.803 & 0.876 \\ \hline
\end{tabular}
\vspace{+1mm}
\caption{The performance of HRNET \cite{Sun2019DeepHR} on the $\frac{1}{2}$ scale dataset and that same dataset upscaled by a factor of 4 using the various SR techniques. The best result for each evaluation metric is higlighted in bold.}
\label{end_to_end_scale2}
\end{table}

\begin{table}[!htb]
\centering
\begin{tabular}{|c|cllccc|}
\hline
Dataset & AP             & $\text{AP}_\text{M}$ & $\text{AP}_\text{L}$ & AR             & $\text{AR}_\text{M}$ & $\text{AR}_\text{L}$ \\ \hline
COCO $\frac{1}{4}$ Scale & 0.519 & 0.785 & 0.785          & 0.567 & 0.833 & 0.888          \\
Bicubic                  & 0.579 & 0.791 & 0.799          & 0.627 & 0.836 & 0.879          \\
ESRGAN                   & 0.602 & 0.801 & 0.798          & 0.649 & 0.843 & 0.877          \\
ESRNET  & \textbf{0.630} & \textbf{0.812}  & 0.807                & \textbf{0.676} & \textbf{0.856}       & 0.886                \\
USRGAN                   & 0.613 & 0.808 & 0.808          & 0.661 & 0.851 & 0.886          \\
USRNET                   & 0.629 & 0.811 & \textbf{0.817} & 0.675 & 0.852 & \textbf{0.893} \\ \hline
\end{tabular}
\vspace{+1mm}
\caption{The performance of HRNET \cite{Sun2019DeepHR} on the $\frac{1}{4}$ scale dataset and that same dataset upscaled by a factor of 4 using the various SR techniques. The best result for each evaluation metric is higlighted in bold.}
\label{end_to_end_scale4}
\end{table}

Our results show a clear overall improvement (AP and AR) for keypoint detection when evaluating SR instead of low resolution imagery. What is surprising however is that there is still a performance decrease for the larger people ($\text{AP}_\text{L}$) in our $\frac{1}{2}$ scale dataset when SR is applied. As shown in Table \ref{Detection_Results_Scale2}, our object detection results for our $\text{AP}_{\text{L}}$ subgroup was either the same or improved once SR was applied. The results in Table \ref{end_to_end_scale2} however, shows that even though the object detection results have slightly improved, SR has made it harder for our keypoint detector to perform optimally. In other words, it performed worse even with better bounding boxes. As our final contribution for this study we therefore decided to introduce an end-to-end top-down HPE approach that would address this problem. 

\subsection{Mask-RCNN with a Segmentation Area Threshold} 

By first applying SR to an image, Mask-RCNN \cite{maskrcnn} would be used to find both the bounding box and segmentation area of people within each image. If the initial segmentation area of a particular person was above a given threshold, then keypoint detection is performed on the original image in a re-scaled bounding box. If the area was below the threshold however, then the SR image is used throughout the end-to-end HPE process. By using one of the best performing SR approaches (USRNET) and the $\frac{1}{2}$ scale dataset, we performed end-to-end top-down HPE using a segmentation area threshold to decide if the SR or original image should be used during the keypoint detection step. Our threshold chosen was a segmentation area of 3500 or less in the original image, as this is where we began to observe minimal benefits from SR as shown in Figure \ref{fig:keypoint_across_res_scale2}. If the persons segmentation area was below this value then the SR image would be used during keypoint detection. If their segmentation area was greater however, then the original image would be used during keypoint detection instead. The results of our mixed approach when compared to simply using the $\frac{1}{2}$ scale and USRNET SR dataset can be seen in Table \ref{mask_rcnn_results}.

\begin{table}[!htb]
\centering
\begin{tabular}{|c|cllccc|}
\hline
Dataset              & AP             & $\text{AP}_\text{M}$ & $\text{AP}_\text{L}$ & AR             & $\text{AR}_\text{M}$ & $\text{AR}_\text{L}$ \\ \hline
COCO $\frac{1}{2}$ Scale & 0.704 & 0.758 & \textbf{0.835} & 0.747 & 0.799 & 0.879 \\
USRNET                   & 0.722 & 0.766 & 0.828          & 0.766 & 0.803 & 0.876 \\
Mixed Approach (W/Threshold) & \textbf{0.723} & \textbf{0.767}       & \textbf{0.835}       & \textbf{0.768} & \textbf{0.804}       & \textbf{0.882}       \\ \hline
\end{tabular}
\caption{The keypoint detection results from HRNET \cite{Sun2019DeepHR} on the $\frac{1}{2}$ scale, USRNET and mixed approach datasets.}
\label{mask_rcnn_results}
\end{table}

As the people in the large subgroup all have a segmentation area above the threshold, the $\text{AP}_\text{L}$ of the original $\frac{1}{2}$ scale dataset and mixed approach are now identical. The $\text{AR}_\text{L}$ has improved however, and this is due to the our object detector finding more large people in the SR than in the low resolution image, as shown by the increase in $\text{AP}_\text{L}$ and $\text{AR}_\text{L}$ for USRNET in Table \ref{Detection_Results_Scale2}. Overall the threshold approach allowed our keypoint detector to perform at its optimum for every evaluation metric, showing that our approach of using Mask-RCNN with a threshold may be a suitable solution for situations where people are both high and low resolution in the same image.


\section{Conclusion}

In this paper we undertook a rigorous empirical study to understand how SR affects the different stages of a top-down HPE process. Prior studies, as well as our initial object detection results, lead us to believe that our final HPE results would also improve once SR was applied; however, this was not the case. Figure \ref{fig:keypoint_across_res_scale2} shows a clear downward trend, showing that as the initial segmentation area of an object increases, the keypoint detection results after SR decreases. {Additionally, as current state of the art keypoint detectors share a similar feed forward architecture and use the same loss function we see no reason why this observation would differ for a different model, simply the threshold at which the performance decreased would change.} Strangely though, our object detector did not seem to exhibit the same downward pattern; instead the change in object detection rate became sporadic for our larger segmentation area subgroups once SR was applied. {This shows that although both components of a top-down HPE model are reliant on an images resolution to perform optimally, the keypoint detection component relies more on this factor than the object detector, whose performance may be more affected by things such as an images context, the lighting of the people in an image, whether people are occluded and if they have a difficult to identify bounding box. Our reasoning for why each components performance degrades as the initial segmentation area increases, is due to the training data that SR models use. Both SR models \cite{USRNET} \cite{ESRGAN} were trained to reconstruct high resolution imagery from their low resolution counterparts \cite{div2k}, meaning that they have not been trained to reconstruct even higher resolution images from medium resolution counterparts. As we increase the segmentation area of the person we wish to reconstruct, they become increasingly higher in resolution. This causes our SR models to struggle as they has not learnt how to deal with inputs of this size.} Although we presented a way to address this problem via our Mask-RCNN approach, the bias introduced by our threshold, as well as not addressing the sporadicity in object detection makes this a sub-optimal solution. {Furthermore, our solution only addresses the issue in top-down HPE approaches as it utilises an object detector therefore we would be unable to apply our solution for bottom-up HPE.} Future works to find an optimum solution could include an end-to-end HPE model which would learn where in an image to apply SR to, as well as a SR approach which could perform optimally on both low and high resolution objects. Overall however, the improvement in HPE when evaluating the effects of SR on low resolution people is noteworthy, and suggests that SR could be used as a valuable tool for future HPE applications in low resolution scenarios.
\bibliography{egbib}
\end{document}